# SOLUTION OF THE DECISION MAKING PROBLEMS USING FUZZY SOFT RELATIONS

**Arindam Chaudhuri**
Lecturer
Mathematics & Computer Science
Meghnad Saha Institute of Technology
Kolkata, India
Email: arindam_chau @ yahoo.co.in

**Kajal De**
Professor
Department of Mathematics
School of Science
Netaji Subhas Open University
Kolkata, India
Email: kajalde@rediffmail.com

**Dipak Chatterjee**
Distinguished Professor
Department of Mathematics
Faculty of Science
St. Xavier's College
Kolkata, India

**Abstract**

The Fuzzy Modeling has been applied in a wide variety of fields such as Engineering and Management Sciences and Social Sciences to solve a number Decision Making Problems which involve impreciseness, uncertainty and vagueness in data. In particular, applications of this Modeling technique in Decision Making Problems have remarkable significance. These problems have been tackled using various theories such as Probability theory, Fuzzy Set Theory, Rough Set Theory, Vague Set Theory, Approximate Reasoning Theory etc. which lack in parameterization of the tools due to which they could not be applied successfully to such problems. The concept of Soft Set has a promising potential for giving an optimal solution for these problems. With the motivation of this new concept, in this paper we define the concepts of Soft Relation and Fuzzy Soft Relation and then apply them to solve a number of Decision Making Problems. The advantages of Fuzzy Soft Relation compared to other paradigms are discussed. To the best of our knowledge this is the first work on the application of Fuzzy Soft Relation to the Decision Making Problems.

## I. Introduction

In today's fast moving world the need for sound, rational Decision Making by Business, Industry and Government is vividly and sometimes disquietingly apparent. A decision is the selection from two or more courses of action. Decision Making can be regarded as an outcome of mental processes which are basically cognitive in nature leading to the selection of a *course of action* among several alternatives. Every Decision Making process produces a final choice [12]. The output can be an *action* or an *opinion of choice*. Decision Making is vital for all categories of problems which may be either long-range or short-range in nature; or the problem may be at relatively high or low level managerial responsibility. The Decision Theory provides a rich set of concepts and techniques to aid the Decision Maker in dealing with complex decision problems The general Decision Theory is defined as follows [18]:

1. *A process which results in the selection from a set of alternative courses of action, that course of action which is considered to meet the objectives of the decision problem more satisfactorily than others as judged by the decision maker.*

2. *The process of logical and quantitative analysis of all factors that influences the decision problem, assists the decision maker in analyzing these problems with several courses of action and consequences.*

The inherent analysis in the Decision Theory is a discipline providing various tools for modeling decision situations in view of explaining them or prescribing actions increasing the coherence between the possibilities offered by the situation, and goals and values systems of agents involved. Mathematical Decision Analysis consists in building a functional or a relational model. The human performance in Decision Making terms has been subject of active research from several perspectives. From a psychological perspective, it is necessary to examine individual decisions in the context of a set of needs, preferences an individual has and values he seeks. From a cognitive perspective, the Decision Making process must be regarded as a continuous process integrated in the interaction with the environment. From a normative perspective, the analysis of individual decisions is concerned with the logic of Decision Making and rationality and the invariant choice it leads to [17]. At another level, it might be regarded as a problem solving activity which is terminated when a satisfactory solution is found. Therefore, Decision Making is a reasoning or emotional process which can be rational or irrational, can be based on explicit assumptions or tacit assumptions.

Logical Decision Making is an important part of all technical professions, where specialists apply their knowledge in a given area to making informed decisions. Some research using natural methods shows, that in situations with higher time pressure, higher stakes, or increased ambiguities, experts use intuitive Decision Making rather than structured approaches, following a recognition primed decision approach to fit a set of indicators into the expert's experience and immediately arrive at a satisfactory course of action without weighing alternatives. Also, recent robust decision efforts have formally integrated uncertainty into the Decision Making process. The role of human judgment and the factors associated with the *fallibility* of Decision Making have been central facets in many areas of human performance research [11]. Attempts to understand Decision Making have generated a rich history of psychological research, much of which is characterized by building formal mathematical and computational models. These models have been used for a variety of purposes, across a range of disciplines and settings. For example, in research on Artificial Intelligence, researchers have studied intelligent agents acting in environments and concerned themselves with the decisions that agents should make. The alternative situation has also been of interest. Researchers observe human agents acting in an environment and attempt to model why certain decisions are made. Such efforts focus on the *policies* that individuals are presumed to be using.

The inherent feature revolving all Decision Making Problems is the vagueness or uncertainty aspects. In order to tackle this problem most psychological researchers, make use of probability. However, two potential issues arise with using probabilistic models. First, some natural sources of uncertainty may not exist in a form that fits a known probability distribution. Second, for modeling cognitive phenomena, the abstract or subjective nature of many cognitive processes may reflect a type of uncertainty that is not conceptually congruent with probability theory and randomness [14].The basic idea that conventional mathematics should be augmented to describe complex systems prompted Lotfi Zadeh to develop the theory of Fuzzy Sets [33] and later generalized into Soft Computing encapsulating techniques such as Fuzzy Systems, Neural Networks, and Genetic Algorithms [34]. Fuzzy Set Theory and Fuzzy Logic provide a system of mathematics that map directly into natural language, thus capturing complex interactions between variables in qualitative descriptions that lend themselves to everyday reasoning. The potential of the Fuzzy System approach for modeling human judgment and decision making lies in several critical features such as model free estimators or universal approximations, imprecision associated with everyday reasoning and the representation of human judgment models as fuzzy rules.

Biases can creep into our Decision Making processes. Many different people have made a decision about the same question and then craft potential cognitive interventions aimed at improving Decision Making outcomes. Some the commonly argued cognitive biases are selective search for evidence; premature termination of search for evidence; inertia; selective perception; wishful thinking or optimism bias; choice supportive bias; Recency; repetition bias; anchoring and adjustment; source credibility bias; incremental Decision Making; attribution symmetry; role fulfillment; underestimating uncertainty and the illusion of control.

Some of the Decision Making techniques that we use in everyday life include listing the advantages and disadvantages of each option commonly used by Plato and Benjamin Franklin; flipping a coin, cutting a deck of playing cards, and other random or coincidence methods; accepting the first option that seems like it might achieve the desired result; prayer, tarot cards, astrology, augurs, revelation, or other forms of divination; acquiesce to a person in authority or an *expert*; calculating the expected value or utility for each option. Let us consider an example. A person is considering two jobs. At the first job option the person has a 60% chance of getting a 30% raise in the first year. And at the second job option the person has an 80% chance of getting a 10% raise in the first year. The decision maker would calculate the expected value of each option, calculating the probability multiplied by the increase of value. As such the expected value for option $a = 0.60 * 0.30 = 0.18$ and for option $b = 0.80 * 0.10 = 0.08$. The person deciding on the job would choose the option with the highest expected value, in this example option number one.

In this paper, we devise optimal solutions for many complex problems in the Engineering, Management and Social Science disciplines which involve data that are not always precisely defined using the concept of Fuzzy Soft relation. The Fuzzy Soft relation has its origins in the Soft Sets which was initially given by Molodtsov [25]. The Decision Making Problems considered here are House Acquisition Problem, Job Allocation Problem, Investment Portfolio Problem, Fund Sources Problem, Manpower Recruitment Problem, and Product Marketing Problem. These problems have various types of uncertainties, some of which can be dealt with using theories viz., Probability Theory, Fuzzy Set Theory [33], Rough Set Theory [28], Vague Set Theory and Approximate Reasoning Theory. However, all these techniques lack in parameterization of the tools due to which these could not be applied successfully in tackling such problems. The Soft Sets concept is free from above difficulty, and has a rich potential for application for these problems. With the motivation of this new concept, we define Soft Relation and Fuzzy Soft relation, which are certain extensions of crisp and fuzzy relations respectively and apply them to solve the above Decision Making Problems.

Rest of this paper is organized as follows. In section 2, the general Decision Making problem is discussed. This is followed by some work related to the Decision Making Problems in section 3. In the next two sections, concepts of the Soft Relation and Fuzzy Soft Relation are illustrated. In section 6, an alternative approach to the Soft Relation and Fuzzy Soft Relation are presented. Various real life applications of Fuzzy Soft Relation follow in the next section. Finally, in section 8 conclusions are given.

## II. Decision Making Problem

Decision Theory or Decision Analysis [18] can be used to determine optimal strategies where a decision maker is faced with several decision alternatives and an uncertain pattern of future events. For example, a manufacturer of a new style of clothing would like to manufacture large quantities of the product if the consumer acceptance and consequently demand for the product are going to be high. Likewise, the manufacturer would like to produce smaller quantities if the consumer acceptance and demand for the product are going to be low. Unfortunately, seasonal clothing items require the manufacturer to make a production quantity decision before the demand is actually known. The actual consumer acceptance of the new product will not be determined until the items have been placed in the stores and the consumers have had the opportunity to purchase them. The selection of the best production volume decision from among several production volume alternatives when the decision maker is faced with the uncertainty of future demand is a problem for the Decision Theory Analysis. Decision Theory commences with the assumption that regardless of the type of decision involved, all the Decision Making Problems have certain common characteristics which are briefly enumerated below:

1. *The Decision Maker*: The *decision maker* refers to individual or a group of individuals responsible for making the choice of an appropriate course of action amongst the available course of action.

2. *Courses of Action*: The *courses of action* or *strategies* are the acts that are available to the *decision maker*. The Decision Analysis involves a selection among two or more *courses of action* and the problem is to choose the best of these alternatives, in order to achieve an objective.

3. *States of Nature*: The events identify the occurrences which are outside of the *decision maker's* control and which determine the level of success for a given act. These events are often called *states of nature* or *outcomes*.

4. *Payoff*: Each combination of a *course of action* and a *state of nature* is associated with a *payoff*, which measures the net benefit to the *decision maker* that accrues from a given combination of decision alternatives and events.

5. *Payoff Table*: For a given problem, *payoff table* lists the *states of nature* which are mutually exclusive as well as collectively exhaustive and a set of given *courses of action* or s*trategies*. For each combination of *states of nature* and *course of action*, the *payoff* is calculated. Suppose the problem under consideration has *m* possible *events* or *states of nature* denoted by $S_1,\ldots,S_m$ and *n courses of action* denoted by $A_1,\ldots,A_n$. Then the *payoff* corresponding to strategy $A_j$ of the *decision maker* under the *state of* nature $S_i$ will be denoted by $p_{ij}$ ($i = 1,……..,m; j = 1,……….,n$). The *mn payoff* can be conveniently arranged in a tabular form known as *m × n payoff table*.

| States of Nature | Conditional Payoffs *Courses of Action* (*Strategies*) | | | | | |
|---|---|---|---|---|---|---|
| | $A_1$ | $A_2$ | ………………. | | | $A_n$ |
| $S_1$ | $p_{11}$ | $p_{12}$ | ……. | …….. | …….. | $p_{1n}$ |
| $S_2$ | $p_{21}$ | $p_{22}$ | ……. | …….. | …….. | $p_{2n}$ |
| ……. | …….. | …….. | ……. | …….. | …….. | …….. |
| ……. | …….. | …….. | ……. | …….. | …….. | …….. |
| $S_m$ | $p_{m1}$ | $p_{m2}$ | ……. | …….. | …….. | $p_{mn}$ |

**Table 1: General Form of Payoff Table**

6. *Regret or Opportunity Loss Table*: The *opportunity loss* is defined as the difference between the possible profits for a *state of nature* and the actual profit obtained for the particular action taken. *Opportunity losses* are calculated separately for each *state of nature* that might occur. Consider a fixed *state of nature* $S_i$. The payoffs corresponding to the $n$ strategies are given by $p_{i1},……..,p_{in}$. Suppose $M_i$ is the maximum of these quantities. Then if $A_1$ is used by the decision maker there is *loss of opportunity* of $M_1 - p_{11}$. The table showing the *opportunity loss* can be computed as follows:

| States of Nature | Conditional Opportunity Loss *Courses of Action* (*Strategies*) | | | | | |
|---|---|---|---|---|---|---|
| | $A_1$ | $A_2$ | ………………. | | | $A_n$ |
| $S_1$ | $M_1 - p_{11}$ | $M_1 - p_{12}$ | ……. | …….. | …….. | $M_1 - p_{1n}$ |
| $S_2$ | $M_2 - p_{21}$ | $M_2 - p_{22}$ | ……. | …….. | …….. | $M_2 - p_{2n}$ |
| ……. | …….. | …….. | ……. | …….. | …….. | …….. |
| ……. | …….. | …….. | ……. | …….. | …….. | …….. |
| $S_m$ | $M_m - p_{m1}$ | $M_m - p_{m2}$ | ……. | …….. | …….. | $M_m - p_{mn}$ |

**Table 2: General form of Regret Table**

## III. Related Work

The general Decision Making problem is a NP Complete problem with applications in wide range of disciplines. The problem was initially studied in the mid of $20^{th}$ century by a number of researchers. In 1947 Brunswik [5] gave the foundational work on the *lens model*, where policy capturing denotes a methodology for studying individual differences in decision strategies via mathematical or statistical models. Policy capturing has been employed to study a range of decision environments. In this approach, a set of judgment stimuli, created on the basis of manipulated *cues*, are presented to participants so that their ensuing judgments can be captured and subsequently modeled. Internal validity is addressed through systematic manipulation of the environmental cues, whereas external validity is addressed through the use of expert decision makers. Brehmer and Brehmer [4] addressed certain fundamental issues in using such an approach that include the degree to which individuals use different decision policies and the actual awareness of the strategies they use. In addition, this type of modeling research can play a pivotal role in understanding how to train individuals to use a given policy. Gobet and Ritter [8] gave the judgment modeling research of this type that capitalizes on the advantages of individual level data analysis.

Although the potential benefits of modeling Decision Making are numerous a review of traditional modeling approaches e.g., linear regression reveals a number of factors suggesting that research on alternative methods is warranted. These factors include the use of unrealistic, orthogonal *judgment cues*, arising from the difficulty in analyzing *inter-correlated cues* with multiple regression [3]; reliance on linear models even though the cited *pervasiveness of linearity* may be more reflective of a lack of research on alternative models [7]; and limited selection of methods for eliciting participants verbal descriptions of their judgment policies.

According to behaviorist Isabel Briggs Myers [24], a person's Decision Making process depends on a significant degree on their cognitive style. Myers developed a set of four bipolar dimensions, called the Myers Briggs Type Indicator (MBTI). The terminal points on these dimensions are: *thinking* and *feeling*; *extroversion* and *introversion*; *judgment* and *perception*; and *sensing* and *intuition*. She claimed that a person's Decision Making style is based largely on how they score on these four dimensions. Other studies suggest that these national or cross cultural differences exist across entire societies. For instance Maris Martinsons has found that American, Japanese and Chinese business leaders each exhibit a distinctive national style of Decision Making [23].

Zadeh [33] conceived the concept of Fuzzy Sets and later in the idea of Soft Computing [34] to deal with impreciseness and uncertainty involved in all Decision Making Problems. The Fuzzy Set Theory has been a clever disguise of the Probability Theory and is very much suitable modeling any real life phenomena. In 1980s and 1990s a large number of researchers applied Fuzzy Sets and Soft Computing concepts to solve many problems in Engineering and Management. Some notable works in this direction are given by Jang and Sun [15]. A key concept in Fuzzy Systems Theory and related techniques is the idea of adaptive, model-free estimation. Kosko [20] in discussing Soft Computing techniques quoted that the Intelligent Systems adaptively estimate continuous functions from data without specifying mathematically how outputs depend on input. Essentially this statement refers to the ability of fuzzy systems to map an input domain X e.g., *cues* to an output range Y e.g., *judgments/decisions* without denoting the function $f: X \rightarrow Y$. However, it has been demonstrated mathematically that fuzzy systems are *universal approximations* of continuous functions of a rather general class [19]. Because of this distinction as model-free estimators and *universal approximations*, modeling techniques such as Fuzzy Models have an innate freedom from a priori assumption of the type of relationships that may exist between variables. Although *universal approximation* places no theoretical limits on the modeling capabilities of Fuzzy Systems, in practice, how to optimally construct a model for a given data set to achieve the full modeling power of the approach remains an open question [17]. Inherent in the claim that model-free estimation is an advantage is the belief that some relationships of interest to human performance researchers depart from the normally assumed linear form and that for exploratory research, the specific form of such models cannot be predetermined. In support of this idea in Hammond [12] suggested that various cognitive and judgment tasks vary on a continuum from intuitive to analytical and that the respective judgment models may similarly change in nature and complexity. In view of the wealth of evidence accruing in the physical and life sciences Barton [27] said that many real systems function through complex nonlinear interactions where adaptive modeling tools may prove useful to human performance researchers. Similar *universal approximation* methods, such as Neural Networks have been incorporated into the analytical toolbox of researchers interested in modeling elements of human performance.

Craiger and Coovert [6] discussed how Fuzzy Sets can be used to capture linguistic values such as *high* and *low* in variables related to human performance such as job or task experience and performance. In line with these ideas, the variable performance can be captured by specifying a finite universal set that consists of levels of performance using three Fuzzy Sets viz., *high, moderate*, and *low* performance. It is noteworthy that the concept of Fuzzy Set is congruent with early psychometric ideas; for example, pioneers such as L. L. Thurstone [31] put forth the idea that an individual's opinion could be characterized by more than a single point estimate response as suggested by Hesketh and Hesketh [13]. Newell and Simon [26] in their seminal work demonstrated that much of human problem solving could be expressed as *if-then* types of production rules. This finding helped launch the field of Intelligent Systems. Subsequently, Expert and other Intelligent Systems have been implemented to model, capture, and support human Decision Making. However, traditional rule-based systems suffer from several problems, including the fact that human experts are often needed to articulate propositional rules, that the symbolic processing normally used prevents direct application of mathematics and that traditional rule-based systems require a large number of rules that are often *brittle* and thus they are not robust to the often required novel set of data inputs.

## IV. Soft Relation – A classical approach

In this section, we give a brief introduction to the concept of soft relation given by Molodtsov [24], which has the rich potential for application to the Decision Making Problems. We illustrate the concept with an example.

**Definition 1:** A soft relation may be defined as a soft set over the power set of the cartesian product of two crisp sets. If $X$ and $Y$ are two non-empty crisp sets of some Universal set and $E$ is a set of parameters, then a soft relation denoted as $(R, E)$ is defined as a mapping from $E$ to $P(X \otimes Y)$. Let us consider the following example.

**Example 1:** Let $U$ = {Professors teaching in a College}, $M$ = {Male Professors in $U$} = $\{m_1,\ldots\ldots,m_9\}$, $N$ = {Female Professors in $U$} = $\{f_1,\ldots\ldots,f_9\}$. Let $E_1$ and $E_2$ be two sets of parameters given by $E_1$ = {*is father of, is uncle of, is husband of, is grandfather of, is son of, is nephew of*}, $E_2$ = {*is mother of, is aunt of, is wife of, is grandmother of, is daughter of, is niece of*}. Then a soft relation $R$ over $P(M \otimes N)$ corresponding to $E_1$ may be given as $(R, E_1)$ = {$R$ (*is father of*) = {$(m_1, f_1), (m_2, f_3), (m_4, f_6), (m_6, f_7)$}, $R$ (*is uncle of*) = {$(m_2, f_1), (m_3, f_5), (m_5, f_6)$}, $R$ (*is* *husband of*) = {$(m_3, f_1), (m_4, f_7), (m_9, f_6)$}, $R$ (*is grandfather of*) = {$(m_1, f_4), (m_5, f_4), (m_6, f_6)$}, $R$ (*is son of*) = {$(m_1, f_2), (m_1, f_5), (m_4, f_3)$}, $R$ (*is nephew of*) = {$(m_2, f_8), (m_1, f_9), (m_7, f_8)$}}. Another soft relation $R$ over $P(M \otimes N)$ corresponding to $E_2$ may be given as $(R, E_2)$ = {$R$ (*is mother of*) = {$(f_2, m_1), (f_5, m_1), (f_3, m_4)$}, $R$ (*is aunt of*) = {$(f_8, m_2), (f_9, m_1), (f_8, m_7)$}, $R$ (*is wife of*) = {$(f_1, w_3), (f_7, w_4), (f_6, w_9)$}, $R$ (*is grandmother of*) = {$(f_9, m_6), (f_8, m_5), (f_7, m_9)$}, $R$ (*is daughter of*) = {$(f_1, m_1), (f_3, m_2), (f_6, m_4), (f_7, m_6)$}, $R$ (*is niece of*) = {$(f_2, m_2), (f_3, m_3)$}} .

It is evident that $R$ (*is father of*) and $R$ (*is daughter of*) can be derived from each other. Similarly {$R$ (*is husband of*), $R$ (*is wife of*)} and {$R$ (*is son of*), $R$ (*is mother of*)} can also be derived from each other. Again {$R$ (*is nephew of*), $R$ (*is aunt of*)} is derivable relation. There may be many other soft relations over $P(M \otimes N)$. Each approximate value set in the above soft relation $(R, E_1)$ or $(R, E_2)$ can be expressed in the parameterized matrix form as shown in the Table 3.

Considering the sets $E$ and $P(X \otimes Y) = V$, and any subset of the cartesian product $E \times V$ is called a soft binary relation, denoted by $T$. $\forall e \in E, v \in V$, if $\langle e, v \rangle \in T$, then $e$ and $v$ satisfy relation $T$ i.e., $eTv$; otherwise, $\langle e, v \rangle \notin T$, i.e., $e$ and $v$ do not satisfy relation $T$. The corresponding soft binary relation can be represented as a matrix $M_T = (t_{ij})_{n \times m}$ is called the relation matrix of $T$, where

$$t_{ij} = \begin{cases} 1, \langle e_i, v_j \rangle \in T \\ 0, \langle e_i, v_j \rangle \notin T \end{cases} \quad i = 1, \ldots, n; \, j = 1, \ldots, m$$

Let $T$ be a relation on a non-empty set $E$. $\forall a, b, c \in E$, if the relation satisfies: (1) reflexivity i.e., $aTa$; (2) symmetry i.e., $aTb \Rightarrow bTa$, $T$ is called a soft similarity relation. Furthermore, if $T$ satisfies transitivity i.e., $aTb, bTc \Rightarrow aTc$, then $T$ is a soft equivalence relation. Here, soft similarity relations are denoted as $Y$ and soft equivalence relations $Q$.

Given a set $E$ and a soft similarity relation $Y$ on $E$, $\forall e \in E$, a set $[e]_Y$, called soft similarity class of $e$ induced by relation $Y$, where $[e]_Y = \{e_i | \langle e_i, e \rangle \in Y, e_i \in E\}$. $Q$ is a soft equivalence on $E$, correspondingly, $\forall e \in E$, we define a soft equivalence subset of $e$ with respect to relation $Q$, denoted by $[e]_Q$, where $[e]_Q = \{e_i | \langle e_i, e \rangle \in Q, e_i \in E\}$.

Let $E$ be a set, and $\Sigma$ is a family of sets constituting with non-empty subset of $E$, viz., $\Sigma = \{\Sigma_\lambda | \Sigma_\lambda \neq \emptyset, \Sigma_\lambda \subseteq E, \lambda \in \Re\}$, where $\Re$ is a subscript set. If $\forall e \in E$, there is $\lambda_e \in \Re$ so that $e \in \Sigma_{\lambda_e}$, the set family is called a cover of $E$. It is easy to verify that $\cup_{\lambda \in \Re} \Sigma_\lambda = E$ if $\Sigma$ is a cover of $E$. For any $\lambda_1, \lambda_2$ in $\Re$, if $\Sigma_{\lambda_1} \cap \Sigma_{\lambda_2} = \Phi$, then $\Sigma$ is a soft partition of $E$. Given any set $E$, $\Sigma = \{E\}$ is the coarsest partition and $\prod = \{\{e_i\} | e_i \in E\}$ is the finest partition. Let $\Sigma_1 = \{\Sigma_\lambda | \lambda \in M\}$ and $\prod = \{\prod_\gamma | \gamma \in N\}$ be two partitions of set $E$. $\forall \lambda_0 \in M, \exists \gamma_0 \in N$, such that $\Sigma_{\lambda_0} \subseteq \prod_{\gamma_0}$, then the partition $\Sigma$ is finer than $\prod$, denoted by $\Sigma \prec \prod$. For the partitions $\Sigma_1$ and $\prod$ of set $E$, $V = \{\Sigma_\lambda \cap \prod_\gamma | \Sigma_\lambda \in \Sigma, \prod_\gamma \in \prod, \Sigma_\lambda \cap \prod_\lambda \neq \emptyset\}$ is also a soft partition of set $E$. Moreover, $V \prec \Sigma$, $V \prec \prod$.

## V. Fuzzy Soft Relation

Here we discuss the concept of fuzzy soft relation which is certain extensions of the crisp soft relation. The fuzziness aspect deals with uncertainty and vagueness inherent in the Decision Making Problems. The definition of fuzzy soft relation is followed by an example. We further extend this concept on the relation on two fuzzy soft sets and give the Extension Principle.

**Definition 2:** A fuzzy soft relation may be defined as a soft set over the fuzzy power set of the cartesian product of two crisp sets. If $P(X \otimes Y)$ is the fuzzy power set; $X$ and $Y$ are two non-empty crisp sets of some Universal set and $E$ is a set of parameters, then a function $R: E \rightarrow P(X \otimes Y)$ is called a fuzzy soft relation. For each $\varepsilon \in E$ each ordered pair in $R(\varepsilon)$ has a degree of membership in the fuzzy soft relation $R$, indicating the strength of $\varepsilon$-parametric relationship presents between the elements of the ordered pairs in $R$. Let us consider the following example.

| $R$(is husband of) | $m_1$ | $m_2$ | $m_3$ | $m_4$ | $m_5$ | $m_6$ | $m_7$ | $m_8$ | $m_9$ |
|---|---|---|---|---|---|---|---|---|---|
| $f_1$ | 0 | 0 | 1 | 0 | 0 | 0 | 0 | 0 | 0 |
| $f_2$ | 0 | 0 | 0 | 0 | 0 | 0 | 0 | 0 | 0 |
| $f_3$ | 0 | 0 | 0 | 0 | 0 | 0 | 0 | 0 | 0 |
| $f_4$ | 0 | 0 | 0 | 0 | 0 | 0 | 0 | 0 | 0 |
| $f_5$ | 0 | 0 | 0 | 0 | 0 | 0 | 0 | 0 | 0 |
| $f_6$ | 0 | 0 | 0 | 0 | 0 | 0 | 0 | 0 | 1 |
| $f_7$ | 0 | 0 | 0 | 1 | 0 | 0 | 0 | 0 | 0 |
| $f_8$ | 0 | 0 | 0 | 0 | 0 | 0 | 0 | 0 | 0 |
| $f_9$ | 0 | 0 | 0 | 0 | 0 | 0 | 0 | 0 | 0 |

**Table 3: Parameterized matrix for the soft relation *is husband of***

**Example 2:** Let $P$ = {Paris, Berlin, Amsterdam} and $Q$ = {Rome, Madrid, Lisbon} be two set of cities, and $E$ = {*far*, *very far*, *near*, *very near*, *crowded*, *well managed*}. Let $R$ be the fuzzy soft relation over the sets $P$ and $Q$ given by $(R, E)$ = {$R$ (*far*) = {(Paris, Rome)/ 0.60, (Paris, Madrid)/ 0.45, (Paris, Lisbon)/ 0.40, (Berlin, Rome)/ 0.55, (Berlin, Madrid)/ 0.65, (Berlin, Lisbon)/ 0.70, (Amsterdam, Rome)/ 0.75, (Amsterdam, Madrid)/ 0.50, (Amsterdam, Lisbon)/ 0.80}}. This information can be represented in the form of two-dimensional array (matrix) as shown in the Table 4. It is obvious from the matrix given in Table 4 that a fuzzy soft relation may be considered as a parameterized fuzzy relation.

The soft relation $T$ defined over the non-empty crisp sets takes either of the two values $\{0, 1\}$. In this case the relation matrix is basically a Boolean matrix. In the case of fuzzy soft relation, the values are considered in the interval $[0, 1]$, viz., $t \in [0, 1]$ where the grades of the relations signify the strength that the elements satisfy the relation. In the context of fuzzy soft relation, the properties of fuzzy relations are defined as: (1) reflexivity i.e., $T(a, a) = 1$; (2) symmetry i.e., $T(a, b) = T(b, a)$; (3) transitivity i.e., $T(a, c) \geq MAX_c (T(a, b) \wedge T(b, c))$, where $\wedge$ is a $t$-norm. Here, we consider $min$ as the $t$-norm. The relation is called a fuzzy soft similarity relation if it satisfies the conditions of reflexivity and symmetry. The relation is called a $T$-indistinguishability relation or a fuzzy soft equivalence relation if a soft similarity relation satisfies the $t$-transitivity. Generally, fuzzy soft equivalence relations are also called soft similarity relations. Usually, the concepts of soft similarity relation, fuzzy soft similarity relation, soft equivalence relation and fuzzy soft equivalence relation are used for distinguishing among objects. If $T_1$ and $T_2$ are two soft fuzzy relations on set $E$, the following operators can be defined:

(1) Union: $(T_1 \cup T_2)(a, b) = max \{T_1(a, b), T_2(a, b)\}, \forall a, b \in E$;
(2) Intersection: $(T_1 \cap T_2)(a, b) = min \{T_1(a, b), T_2(a, b)\}, \forall a, b \in E$;
(3) Containment: $T_1 \subseteq T_2 \Rightarrow T_1(a, b) \leq T_2(a, b), \forall a, b \in E$.

| $R$ (*far*) | Rome | Madrid | Lisbon |
|---|---|---|---|
| Paris | 0.60 | 0.45 | 0.40 |
| Berlin | 0.55 | 0.65 | 0.70 |
| Amsterdam | 0.75 | 0.50 | 0.80 |

**Table 4: Parameterized matrix for the fuzzy soft relation *far***

Given a fuzzy relation $T$ on $E$, $\forall \alpha \in [0, 1]$, the $\alpha$-cuts $T_\alpha$ of the fuzzy soft relation is a crisp soft relation, where $T_\alpha (a, b) = \begin{cases} 1, T(a,b) \geq \alpha \\ 0, T(a,b) \leq \alpha \end{cases}$; $T$ is a fuzzy soft equivalence relation if and only if the $\alpha$-cuts $T_\alpha$ of $T$ is a crisp soft equivalence relation for all $\alpha \in [0, 1]$. Given a finite set $E$ and a fuzzy soft equivalence relation $T$, the fuzzy soft equivalence class $[e_i]_T$ of $e_i \in E$ is a fuzzy subset, where $[e_i]_T$ is defined as $[e_i]_T = \frac{t_{i1}}{e_1} + \frac{t_{i2}}{e_2} + \ldots\ldots\ldots + \frac{t_{in}}{e_n}$ where, $t_{ij} = T(e_i, e)$. The fuzzy soft equivalence class is a fuzzy information granule, whereby the elements in the class are fuzzy indiscernible with $e_i$; $t_{ij}$ means the degree how the two elements are equivalent or indiscernible. The family of the fuzzy soft equivalence classes $[e_i]_T$, written as $E/T = \{[e_i]_T \mid e_i \in T\}$, is called a fuzzy soft quotient set of $E$ induced by $T$.

Now, we present the novel uncertainty measures for fuzzy soft binary relations, where we consider the case without probability distributions.

**Definition 3:** Given a finite set $E$ and a fuzzy soft binary relation $T$ on $E$, we define a soft fuzzy class $[e_i]_T$ of $e_i \in E$ induced by the relation $T$, where $[e_i]_T = \frac{t_{i1}}{e_1} + \frac{t_{i2}}{e_2} + \ldots + \frac{t_{in}}{e_n}$. $[e_i]_T$ is a fuzzy soft set, and the fuzzy cardinal number of $[e_i]_T$ is defined as $|[e_i]_T| = \sum_{j=1}^{n} t_{ij}$. For a finite set $E$, $\forall e_j \in E$ we have $t_{ij} \leq 1$, then the cardinality of $[e_i]_T$ is also finite and $|[e_i]_T| \leq n$.

**Definition 4:** Let $E$ be a finite set and $T$ a fuzzy soft relation on $E$. The fuzzy soft relation class of $e$ is $[e_i]_T$, then, the expected cardinality of $[e_i]_T$ is computed as $\overline{card}([e_i]_T) = \frac{|[e_i]_T|}{|E|} = \frac{\sum_{j=1}^{n} t_{ij}}{n}$

The expected cardinality $\overline{card}([e_i]_T) \leq 1$, can be considered as the ratio of $[e_i]_T$ in $E$.

**Definition 5:** The uncertainty quantity of the fuzzy soft relation class $[e_i]_T$ is defined as $V([e_i]_T) = -\log_2 \overline{card}([e_i]_T)$. As $\overline{card}([e_i]_T) \leq 1$, we have $V([e_i]_T) \geq 0$ and, $V([e_i]_T)$ decreases monotonously with the increase of $\overline{card}([e_i]_T)$.

**Definition 6:** Given a finite set $E$ and a fuzzy soft relation $T$ on $E$, we calculate the average uncertainty quantity $G(T)$ of the fuzzy soft relation with $G(T) = -\sum_{i=1}^{n} \frac{1}{n} \log_2 \overline{card}([e_i]_T)$. The average uncertainty quantity of the fuzzy soft relation $T$ on $E$ is a mapping $G : (E,T) \to \Re^+$, where $\Re^+$ is the domain of nonnegative real numbers. With this mapping we form an order to compare the fuzzy soft relations with respect to the uncertainty quantity. It is to be noted that the uncertainty quantity is not only a function of the fuzzy soft relation $T$, but also related to the set $E$. Also, we define $G(T) = 0$ if $E = \emptyset$.

**Proposition 1:** *Given a non-empty and finite set E and a soft relation T on E, if $\forall a, b \in E$; $T(a, b) = 1$, then we have $G(T) = 0$.*

**Proposition 2:** *Let $T_1$ and $T_2$ be two fuzzy soft relations on a nonempty and finite set E, we have $T_1 \subseteq T_2 \Rightarrow G(T_2) \geq G(T_1)$.*

**Proposition 3:** *Let $T_1$ and $T_2$ be two fuzzy soft relations on a nonempty and finite set E, we have $G(T_1 \cap T_2) \geq max(G(T_1), G(T_2))$; $G(T_1 \cup T_2) \geq min(G(T_1), G(T_2))$.*

### A. Relation on two Fuzzy Soft Sets

**Definition 7:** Let $(F, A)$ and $(G, B)$ be two fuzzy soft sets over a common Universal set. Then a relation R of $(F, A)$ on $(G, B)$ may be defined as a mapping $R: A \times B \to P(U^2)$ such that for each $e_i \in A$, $e_j \in B$ and for all $u_p \in F(e_i)$, $u_q \in G(e_j)$, the relation $R$ is characterized by the following membership function, $\mu_R(u_l, u_k) = \mu_{F(e_i)}(u_l) \times \mu_{G(e_j)}(u_k)$, where, $u_l \in F(e_i), u_k \in G(e_j)$.

Thus the above mapping is well defined. Higher the value of the membership grade in the relation R for a pair, stronger is the parametric character present between the pair. We consider an example to illustrate this fact.

**Example 3:** Let $U = \{w_1, w_2, w_3, w_4, w_5, w_6\}$ be the set of watches and $A = \{cheap, costly\}$, $B = \{beautiful, in\ a\ golden\ locket\}$. Let $(F, A)$ and $(G, B)$ be two soft sets given by $F\ (cheap) = \{w_1/ 0.1, w_2/ 0.25, w_3/ 0.2, w_4/ 0.6, w_5/ 0.15, w_6/ 0.35\}$, $F\ (costly) = \{w_1/ 1, w_2/ 0.75, w_3/ 0.8, w_4/ 0.55, w_5/ 0.9, w_6/ 0.85\}$, $G\ (beautiful) = \{w_1/ 0.65, w_2/ 1, w_3/ 0.8, w_4/ 0.7, w_5/ 0.8, w_6/ 0.75\}$, $G\ (in\ a\ golden\ locket) = \{w_1/ 0.6, w_2/ 0.75, w_3/ 0.8, w_4/ 0.5, w_5/ 0.45, w_6/ 0.95\}$. The relation $R: A \times B \to P(U^2)$ is given by the following membership matrices in Table 5 and 6. The membership function of $R$ can also be defined using other appropriate techniques.

### B. Extension Principle on Fuzzy Soft Sets

Let X be a cartesian product of the Universes, $X_1, \ldots, X_r$; and $(F, A_1), \ldots, (F, A_r)$ be $r$ fuzzy soft sets in $X_1, \ldots, X_r$. Further consider $f$ to be a mapping from $X$ to $Y$, given by $y = f(x_1, \ldots, x_r)$, then fuzzy soft set B can be defined using extension principle as B = $\{\langle y, \mu_B(y) \rangle\ |\ y = f(x_1, \ldots, x_r),\ (x_1, \ldots, x_r) \in X\}$, where $\mu_B(y) = \max_{(x_1, \ldots, x_r) \in f^{-1}y} \min\{\mu_{(F, A_1)}(x_1), \ldots, \mu_{(F, A_r)}(x_r)\}$ if $f^{-1}(y) \neq \Phi$ otherwise $\mu_B(y) = 0$.

| R (costly, beautiful) | $w_1$ | $w_2$ | $w_3$ | $w_4$ | $w_5$ | $w_6$ |
|---|---|---|---|---|---|---|
| $w_1$ | 0.65 | 1 | 0.80 | 0.70 | 0.80 | 0.75 |
| $w_2$ | 0.49 | 0.75 | 0.60 | 0.53 | 0.60 | 0.56 |
| $w_3$ | 0.52 | 0.80 | 0.64 | 0.56 | 0.64 | 0.60 |
| $w_4$ | 0.35 | 0.55 | 0.44 | 0.39 | 0.55 | 0.42 |
| $w_5$ | 0.59 | 0.90 | 0.72 | 0.63 | 0.72 | 0.68 |
| $w_6$ | 0.53 | 0.85 | 0.68 | 0.60 | 0.68 | 0.64 |

Table 5: Membership matrix for the fuzzy soft relation *costly*, *beautiful*

## VI. Soft Relation and Fuzzy Soft Relation – An Alternative Approach

In this section we consider an alternative approach to the concepts of the soft relation and fuzzy soft relation. The soft relation generally uses two-valued logic and as such the propositions may be either true or false, but not both. As a consequence of this, something which is not true is false and vice versa, i.e., the *law of the excluded middle* holds. This is only an approximation to human reasoning which gives rise to the multi-valued logic in fuzzy soft relation. For example, consider the *class of tall men* which does not constitute classes or sets in the usual mathematical sense of these terms. The term *tall* is an elastic property. To define the *class of tall men* as a crisp set, a predicate $P(x)$ is used, where $x$ may be 176 cm, where $x$ is the height of a person and in the figure 176 denotes the *threshold* value. This is an abrupt approximation to the concept *tall*. From an engineering viewpoint, it is likely that the measurement is uncertain, due to source of noise in the equipment. Thus, measurements within the narrow range of $176 \pm \varepsilon$, where $\varepsilon$ is the variation in noise which could fall on either side of the *threshold* randomly. This leads to the concept of *membership grade*, $\mu_A(x)$ which allows finer detail, such that the transition from membership to non-membership is gradual rather than abrupt. The *membership grade* for all members defines a *fuzzy relation* as given in Figure 1. Corresponding to the *membership grade*, there is a *membership function* that relates $x$ to each *membership grade*, $\mu_A(x)$ which is in fact a real number in the closed interval [0, 1].

| $R$ (*cheap*, *beautiful*) | $w_1$ | $w_2$ | $w_3$ | $w_4$ | $w_5$ | $w_6$ |
|---|---|---|---|---|---|---|
| $w_1$ | 0.07 | 0.10 | 0.08 | 0.07 | 0.08 | 0.08 |
| $w_2$ | 0.16 | 0.25 | 0.20 | 0.18 | 0.20 | 0.19 |
| $w_3$ | 0.13 | 0.20 | 0.16 | 0.14 | 0.16 | 0.15 |
| $w_4$ | 0.39 | 0.60 | 0.48 | 0.42 | 0.48 | 0.45 |
| $w_5$ | 0.01 | 0.15 | 0.12 | 0.11 | 0.12 | 0.11 |
| $w_6$ | 0.23 | 0.35 | 0.28 | 0.25 | 0.28 | 0.26 |

**Table 6: Membership matrix for the fuzzy soft relation *cheap*, *beautiful***

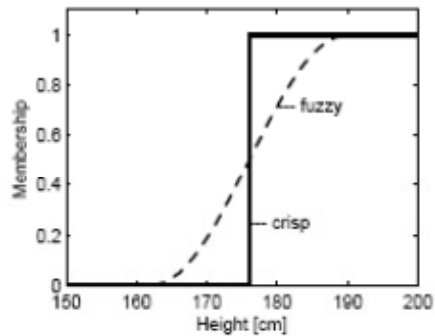

**Figure 1: The definitions term *tall men* in terms of the crisp and fuzzy soft relations.**

The term *fuzzy soft* or *indistinct* suggests an image of a boundary zone, rather than an abrupt frontier. Indeed, the *soft relations* are being considered as relations composed of *crisp sets*, to distinguish them from *fuzzy soft relations*. As with *soft relations*, we are only guided by intuition in deciding which objects are members and which are not; a formal basis for how to determine the *membership grade* of a *fuzzy soft relation* is absent. The *membership grade* is a precise, but arbitrary measure as it rests on personal opinion, not reason. The range of values of *membership grade* is $0 \leq \mu \leq 1$, the higher the value, the higher the *membership grade*. A *soft relation* is consequently a special case of a *fuzzy soft relation*, with membership values restricted to $\mu \in \{0$, 1}. The members of *fuzzy soft relations* are taken from a *universe of discourse* which comprises of all objects that can be taken into consideration and generally depends on the context.

There are two ways to represent a *membership function* viz., continuous or discrete. A continuous *fuzzy soft relation* A is defined by means of a continuous *membership function*, $\mu_A(x)$. A *trapezoidal membership function* is a piecewise linear, continuous function, controlled by four parameters viz., *a*, *b*, *c*, *d* [16]:

$$\mu_{trapezoid}(x;a,b,c,d) = \begin{cases} 0, x \leq a \\ \frac{x-a}{b-a}, a \leq x \leq b \\ 1, b \leq x \leq c \\ \frac{d-x}{d-c}, c \leq x \leq d \\ 0, d \leq x \end{cases} ; x \in \Re$$

The parameters $a \leq b \leq c \leq d$ define the four breakpoints, here designated as: *left footpoint*, *a*; *left shoulderpoint*, *b*; *right shoulderpoint*, *c*; and *right footpoint*, *d* as shown in Figure 2 (a). A *triangular membership function* is piecewise linear, and derived from the *trapezoidal membership function* by merging the two shoulder points into one, i.e., *b* = *c* as shown in Figure 2 (b). Smooth, differentiable versions of the *trapezoidal* and *triangular membership functions* can be obtained by replacing the linear segments corresponding to the intervals $a \leq x \leq b$ and $c \leq x \leq d$ by a nonlinear function, for instance a half period of a cosine function,

$$\mu_{smoothtrapezoid}(x;a,b,c,d) = \begin{cases} 0, x \leq a \\ \frac{1}{2} + \frac{1}{2}\cos(\frac{x-b}{b-a}\pi), a \leq x \leq b \\ 1, b \leq x \leq c \\ \frac{1}{2} + \frac{1}{2}\cos(\frac{x-c}{d-c}\pi), c \leq x \leq d \\ 0, d \leq x \end{cases} ; x \in \Re$$

They are known as *smooth trapezoid* or *soft trapezoid* and *smooth triangular* or *soft triangular* which are illustrated in Figure 2 (c), (d). Other possibilities exist for generating smooth trapezoidal functions, for example Gaussian, generalized bell, and sigmoidal membership functions [16].

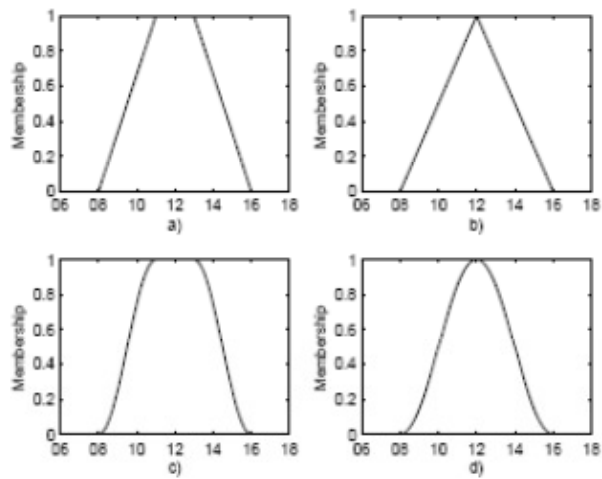

**Figure 2: (a) Trapezoidal Membership Function; (b) Triangular Membership Function; (c) Smooth Trapezoid; (d) Smooth Triangular**

A discrete *fuzzy soft relation* is defined by means of a discrete variable $x_i$ ($i = 1, 2 \ldots$). They are generally defined by ordered pairs, A = $\{\langle x_1, \mu(x_1)\rangle, \langle x_2, \mu(x_2)\rangle \ldots | x_i \in U, i = 1, 2 \ldots\}$. Each membership value $\mu(x_i)$ is an evaluation of the *membership function* $\mu$ at a discrete point $x_i$ in the universe $U$, and the whole set is a collection, usually finite, of pairs $\langle x_i, \mu(x_i)\rangle$.

**Example 4:** To achieve a discrete *triangular membership function* from the *trapezoid membership function* let us assume that the universe is a vector $u$ of 7 elements. In MATLAB notation, $u$ = [9 10 11 12 13 14 15]. Considering the parameters are $a = 10$, $b = 12$, $c = 12$, and $d = 14$ then, by the *trapezoid membership function*, the corresponding membership values are a vector of 7 elements, viz., [0, 0, 0.5, 1, 0.5, 0, 0]. Each membership value corresponds to one element of the universe, more specifically written as given in Table 7, with the universe in the bottom row, and the membership values in the top row. As a crude rule of thumb, the continuous form is more computing intensive, but less storage demanding than the discrete form.

| 0 | 0 | 0.5 | 1 | 0.5 | 0 | 0 |
|---|---|---|---|---|---|---|
| 9 | 10 | 11 | 12 | 13 | 14 | 15 |

**Table 7: Membership values and corresponding elements of the universe**

**Definition 8:** If $(F, A)$ and $(G, B)$ are two soft sets over a common universe $U$, then the soft subset $(R, C)$ of $(F, A) \times (G, B)$ is called a soft relation of $(F, A)$ and $(G, B)$, where $C \subset A \times B$ and for every $(x, y) \in C$, $R(x, y)$ and $S(x, y)$ are identical approximations where $S(x, y) = F(x) \cap G(y)$. It is clear that $(R, C)$ is also a soft set and therefore the basic concepts such as union, intersection, complement, difference and exclusion can be applied without any modification to the soft relation. Let us consider the following example.

**Example 5:** Let $U = \{c_1, c_2, c_3, c_4, c_5, c_6\}$ be the set of cars and $E = \{cheap, costly, fuel efficient, produced by firm A, produced by firm B, produced by firm C\}$ be the set of parameters. Let $(F, P)$ = $\{cheap\ cars = \{c_1, c_2, c_3\}, costly\ cars = \{c_4, c_5\}, fuel\ efficient\ cars = \{c_1, c_3, c_5, c_6\}\}$ and $(G, Q) = \{cars\ produced\ by\ firm\ A = \{c_1, c_3\}, cars\ produced\ by\ firm\ B = \{c_2, c_3, c_4\}, cars\ produced\ by\ firm\ C = \{c_2, c_5, c_6\}\}$ be two soft sets over $U$. Then a soft relation $(R, C)$ of all *cheap* and *fuel efficient* cars *produced by the firms A* and *C* respectively is given by $(R, C) = \{R\ (cheap, produced\ by\ firm\ A) = \{c_1, c_3\}, \{R\ (fuel\ efficient, produced\ by\ firm\ C) = \{c_5, c_6\}$.

Based on the definition 8, we give the generalized representation of the operations viz., AND, OR, NOT, NAND and NOR operations.

a. **AND Operation:** The AND Operation can be generalized for a family of $n$ fuzzy soft sets $\{(F_i, A_i)/ i \in N\}$, denoted by $(F_1, A_1) \wedge (F_2, A_2) \wedge \ldots \ldots \wedge (F_n, A_n) = \underset{i \in N}{\wedge} (F_i, A_i)$ and given by $\wedge (F_i, A_i) = (H, \times A_i) = (H, A_1 \times A_2 \times \ldots \ldots \times A_n)$ where $H(x_1, \ldots \ldots, x_n) = F_1(x_1) \cap F_2(x_2) \cap \ldots \ldots \cap F_n(x_n)\ \forall\ (x_1, \ldots \ldots, x_n) \in A_1 \times \ldots \times A_n$.

b. **OR Operation:** The OR Operation can be generalized for a family of $n$ fuzzy soft sets $\{(F_i, A_i)/ i \in N\}$, denoted by $(F_1, A_1) \vee (F_2, A_2) \vee \ldots \ldots \vee (F_n, A_n) = \underset{i \in N}{\vee}(F_i, A_i)$ and given by $\vee (F_i, A_i) = (P, \times A_i) = (P, A_1 \times A_2 \times \ldots \ldots \times A_n)$ where $P(x_1, \ldots \ldots, x_n) = F_1(x_1) \cup F_2(x_2) \cup \ldots \ldots \ldots \cup F_n(x_n) \; \forall \; (x_1, \ldots \ldots, x_n) \in A_1 \times \ldots \ldots \times A_n$.

c. **NOT Operation:** The NOT Operation can be generalized for a family of $n$ fuzzy soft sets $\{(F_i, A_i)/ i \in N\}$, denoted by $\neg (F_i, A_i)$ and given by $\neg (F_i, A_i) = (P, A_i)$ where $P(x_i) = \neg F_i(x_i) \; \forall \; x_i \in A_i$.

d. **NAND Operation:** The NAND Operation can be generalized for a family of $n$ fuzzy soft sets $\{(F_i, A_i)/ i \in N\}$, denoted by $\neg ((F_1, A_1) \wedge (F_2, A_2) \wedge \ldots \ldots \wedge (F_n, A_n)) = \neg \underset{i \in N}{\wedge}(F_i, A_i)$ and given by $\neg \wedge (F_i, A_i) = (H, \times A_i) = (H, A_1 \times A_2 \times \ldots \ldots \times A_n)$ where $H(x_1, \ldots \ldots, x_n) = \neg F_1(x_1) \cap \neg F_2(x_2) \cap \ldots \ldots \ldots \cap \neg F_n(x_n) \; \forall \; (x_1, \ldots \ldots, x_n) \in A_1 \times \ldots \ldots \times A_n$.

e. **NOR Operation:** The OR Operation can be generalized for a family of $n$ fuzzy soft sets $\{(F_i, A_i)/ i \in N\}$, denoted by $\neg ((F_1, A_1) \vee (F_2, A_2) \vee \ldots \ldots \vee (F_n, A_n)) = \neg \underset{i \in N}{\vee}(F_i, A_i)$ and given by $\neg \vee (F_i, A_i) = (P, \times A_i) = (P, A_1 \times A_2 \times \ldots \ldots \times A_n)$ where $P(x_1, \ldots \ldots, x_n) = \neg F_1(x_1) \cup \neg F_2(x_2) \cup \ldots \ldots \ldots \cup \neg F_n(x_n) \; \forall \; (x_1, \ldots \ldots, x_n) \in A_1 \times \ldots \ldots \times A_n$.

Similarly, the set difference and exclusion operations can also be defined. Considering the above operations we give the following definitions:

**Definition 9:** If $(F_1, A_1), \ldots \ldots, (F_n, A_n)$ be $n$ soft sets then the soft subset $(R, C)$ of $(F_1, A_1) \times (F_2, A_2) \times \ldots \ldots \times (F_n, A_n)$ is called an $n$-ary soft relation. Here, $C \subset A_1 \times A_2 \times \ldots \ldots \times A_n$ and $\forall (x_1, \ldots \ldots, x_n) \in A_1 \times A_2 \times \ldots \ldots \times A_n$, $R(x_1, \ldots \ldots, x_n)$ and $P(x_1, \ldots \ldots, x_n)$ are identical approximations where $P(x_1, \ldots \ldots, x_n) = F_1(x_1) \cap F_2(x_2) \cap \ldots \ldots \ldots \cap F_n(x_n)$.

**Definition 10:** If $(F, A)$ and $(G, B)$ are two fuzzy soft sets then the fuzzy soft subset $(R, C)$ of $(F, A) \times (G, B)$ is called a fuzzy soft relation. Here, $C \subset A \times B$ and $\forall (x, y) \in A \times B$, $R(x, y)$ is a fuzzy subset of $P(x, y)$ where, $P(x, y) = F(x) \cap G(y)$.

**Example 6:** Let $U = \{c_1, c_2, c_3, c_4, c_5, c_6\}$ be the set of cars and $(F, A)$ be a fuzzy soft set which describes *the cost of the cars* and $(G, B)$ be a fuzzy soft set which describes *the attractiveness of the cars* where, $A = \{costly, moderate, cheap\}$ and $B = \{fuel\ efficient, beautiful, having\ metallic\ color\}$. Let, $F(costly) = \{c_1/ 0.5, c_2/ 0.8, c_3/ 0, c_4/ 0.1, c_5/ 1, c_6/ 0.9\}$; $F(moderate) = \{c_1/ 0.2, c_2/ 0.4, c_3/ 0.5, c_4/ 0.6, c_5/ 0.5, c_6/ 0.7\}$; $F(cheap) = \{c_1/ 0.5, c_2/ 0.1, c_3/ 1, c_4/ 0.9, c_5/ 0, c_6/ 0.4\}$; $G(fuel\ efficient) = \{c_1/ 0.4, c_2/ 0.6, c_3/ 0.8, c_4/ 1, c_5/ 0.2, c_6/ 0.5\}$; $G(having\ metallic\ color) = \{c_1/ 1, c_2/ 0, c_3/ 0, c_4/ 1, c_5/ 0, c_6/ 1\}$; $G(beautiful) = \{c_1/ 0.8, c_2/ 0, c_3/ 0.5, c_4/ 0.7, c_5/ 0.9, c_6/ 0.8\}$. Then a fuzzy soft relation $R$ of all *cheap, fuel efficient* and *beautiful cars* is given by $(R, C) = \{R(cheap, fuel\ efficient) = \{c_1/ 0.4, c_2/ 0.1, c_3/ 0.8, c_4/ 0.9, c_5/ 0, c_6/ 0.4\}, R(cheap, beautiful) = \{c_1/ 0.5, c_2/ 0, c_3/ 0.5, c_4/ 0.7, c_5/ 0, c_6/ 0.4\}\}$.

## A. Generalization of *n* Fuzzy Soft Sets

The fuzzy soft relation considered in definition 10 can be generalized to *n* fuzzy soft sets $\{(F_i, A_i) / i \in N\}$ in the following manner.

**Definition 11:** The fuzzy soft set $(R, C)$ of $(F_i, A_i)$ is called an *n*-ary fuzzy soft relation. Here, $C \subset A_1 \times \ldots \ldots \times A_n$ $\forall$ $(x_1, \ldots \ldots, x_n) \in A_1 \times \ldots \ldots \times A_n$, $R(x_1, \ldots \ldots, x_n) \subset O$ where, $O(x_1, \ldots \ldots, x_n) = F_1(x_1) \cap \ldots \ldots \cap F_2(x_2)$.
By analogy the relation on *n*-soft sets is called an *n*-ary or *n*-dimensional relation.

## B. Logic in Fuzzy Soft Sets

The association of logic with crisp and fuzzy soft relations can be considered as the study of language in arguments and persuasion, and is used to judge the correctness of a chain of reasoning in a mathematical proof. The goal is to reduce principles of reasoning to a code. For crisp soft relation the *truth* or *falsity* values are the assigned truth-values of the proposition. The fuzzy soft relation considers the *true* or *false* values of the proposition or an intermediate truth-value such as *maybe true*, which may further be extended to *multi-valued* logic. Generally the unit interval is subdivided into finer divisions to achieve greater level of precision. The logical statements are basically represented as *propositions* or elementary sentences which are combined with *connectives* such as *and* (conjunction), *or* (disjunction), *if-then* (*implies*), *if and only if* (*equivalence*) to form compound propositions. In many practical situations, assertions are used which contains at least one propositional variable called a *propositional form*. The main difference between *proposition* and *propositional form* is that every *proposition* has a truth-value, whereas a *propositional form* is an assertion whose truth-value cannot be determined until propositions are substituted for its propositional variables. But when no confusion results, we refer to *propositional forms* as *propositions*. A *truth-table* summarizes the possible truth-values of an assertion. For example, consider the truth-table for the crisp soft propositional form viz., $p \vee q$. The truth-table in figure 3 lists all possible combinations of truth-values i.e., the cartesian product of the arguments $p$ and $q$ in the two leftmost columns. The rightmost column holds the truth-values of the proposition. Alternatively, the truth-table can be rearranged into a two-dimensional array, also known as the *cayley-table* as shown below.

| p | q | $p \vee q$ |
|---|---|---|
| 0 | 0 | 0 |
| 0 | 1 | 1 |
| 1 | 0 | 1 |
| 1 | 1 | 1 |

|   | 0 | 1 | → q |
|---|---|---|---|
| 0 | 0 | 1 | |
| 1 | 1 | 1 | |
| ↓ p | | | |

**Figure 3: The truth and cayley tables for $p \vee q$**

Along the vertical axis in the cayley table, symbolized by arrow ↓, are the possible values are 0 and 1 of the first argument $p$; along the horizontal axis, symbolized by arrow →, the possible values are 0 and 1 of the second argument $q$. At the intersection of row $i$ and column $j$ is the truth-value of the expression $p_i \vee q_j$. Similarly, the other logical operations can be represented by means of the *truth-table* and *cayley-table*. By analogy, we can define similar truth-tables for *fuzzy soft* logic connectives. We start by defining negation and disjunction; and we can derive the truth-tables of other connectives from that point of departure. Let us define *disjunction* as set union, i.e., $p \vee q \equiv \max(p, q)$. We have the truth-table for the fuzzy soft connective *or* as shown in figure 4.

Like before, the *p*-axis is vertical and the *q*-axis horizontal. At the intersection of row *i* and column *j* the value of the expression is *max* ($p_i$, $q_j$). When looking for definitions of fuzzy soft connectives, we will require that such connectives should agree with their crisp soft counterparts for the truth-domain {0, 1}. In terms of truth-tables, the values in the four corners of the fuzzy soft cayley-table, should agree with the cayley-table for the crisp soft connective. Similarly, other *fuzzy soft* logic connectives can be defined.

$$\begin{array}{c|ccc}
p \vee q & 0 & 0.5 & 1 \\
\hline
0 & 0 & 0.5 & 1 \\
0.5 & 0.5 & 0.5 & 1 \\
1 & 1 & 1 & 1
\end{array}$$

**Figure 4: The truth table for fuzzy soft connective *or***

The *implication* connective however should be taken care of with caution. If we define it as *material implication*, $\neg p \vee q$, then we get a fuzzy soft truth-table which is unsuitable, as it causes several useful logical laws to break down. It is important to realize, that we must make a design choice at this point, in order to proceed with the definition of *implication* and *equivalence*. The choice is which logical laws we wish to apply. Not all laws known from two-valued soft logic can be valid in fuzzy soft logic. Take for instance the propositional form, $p \vee \neg p \Leftrightarrow 1$ which is equivalent to the law of the excluded middle. Testing with the truth-value $p = 0.5$ (fuzzy soft logic) the left hand side yields $0.5 \vee \neg 0.5 = max(0.5, 1 - 0.5) = 0.5$. This is different from the right hand side, and thus the law of the excluded middle is invalid in fuzzy soft logic. If a proposition is true with a truth-value of 1, for any combination of truth-values assigned to the variables, we shall say it is *valid*. Such a proposition is a *tautology*. If the proposition is true for some, but not all combinations, we shall say it is *satisfiable*. One tautology that we definitely wish to apply in fuzzy soft logic applications is $[p \wedge (p \Rightarrow q)] \Rightarrow q$. In other words, if *p* and *p* implies *q* then *q*. The above tautology is closely associated with the *modus ponens* rule of inference. Another tautology that is extensively used is the transitive relationship, $[(p \Rightarrow q) \wedge (q \Rightarrow r)] \Rightarrow (p \Rightarrow r)$. In other words, if *p* implies *q* which in turn implies *r*, then *p* implies *r*. Whether these propositions are valid in fuzzy soft logic depends on how the connectives are defined. Or rather, the connectives are defined, *implication* in particular, such that those propositions become valid. Closely related to the *implication* connective is *inference*. Logic provides principles of reasoning, by means of *inference,* the drawing of conclusions from assertions. The verb *to infer* means to conclude from evidence, deduce, or to have as a logical consequence. *Rules of inference* specify conclusions drawn from assertions known or assumed to be true. One such commonly used rule of inference is *modus ponens*. The generalized form to fuzzy soft logic is the core of fuzzy soft reasoning. It is often presented in the form of the argument given in figure 5.

$$P$$
$$\frac{P \Rightarrow Q}{Q}$$

**Figure 5:** *Modus Ponens* **rule of inference**

In other words, if *P* is known to be true, and $P \Rightarrow Q$ is true, then *Q* must be true. Considering the two-valued soft logic, we see from the cayley-table for *implication* that given in figure 6, whenever $P \Rightarrow Q$ and *P* are true then so is *Q*; by *P* true we consider only the second row, leaving *Q* true as the only possibility. In such an argument the assertions above the line are the *premises*, and the assertion below the line the *conclusion*. It is to be noticed that the premises are assumed to be true, not considering *all* possible truth combinations. On the other hand, underlying *modus ponens* is tautology, which expresses the same, but is valid for *all* truth-values. Therefore modus ponens is valid in fuzzy soft logic, if tautology is valid in fuzzy soft logic.

$$
\begin{array}{c}
p \Rightarrow q \\
\begin{array}{c|cc}
 & 0 & 1 \\
\hline
0 & 1 & 1 \\
1 & 0 & 1
\end{array} \rightarrow q \\
\downarrow \\
p
\end{array}
$$

**Figure 6: The cayley table for fuzzy soft connective** *implication*

The inference mechanism in fuzzy soft *modus ponens* can be generalized. Given a relation *R* connecting logical variables *p* and *q*, we infer the possible values of *q* for a particular instance of *p*; considering the vector-matrix representation, to emphasize the computer implementation, with *p* as column vector and *R* two-dimensional truth-table, with the *p*-axis vertical, the inference is defined as $\mathbf{q}^t = \mathbf{p}^t \circ \mathbf{R}$. The operation ∘ is an inner $\vee - \wedge$ product. The $\wedge$ operation is same as in $p \wedge (p \Rightarrow q)$ and the $\vee$ operation along the columns yields what can possibly be implied about *q*, confer the rightmost implication in $[p \wedge (p \Rightarrow q)] \Rightarrow p$. Assuming *p* is true corresponds to $p = \binom{0}{1}$. But the scheme is more general, because we could also assume *p* is false, compose with **R** and study what can be inferred about *q*. Taking for instance modus ponens, thus $R = \binom{11}{01}$ which is the truth-table for $p \Rightarrow q$. Assigning *p* as above, $\mathbf{q}^t = \mathbf{p}^t \circ \mathbf{R} = (01) \circ \binom{11}{01} = (01)$. The outcome $\mathbf{q}^t$ is a truth-vector pointing at *q* true as the only possible conclusion, as expected. For instance with $p = (10)^t$ yields $\mathbf{q}^t = \mathbf{p}^t \circ \mathbf{R} = (10) \circ \binom{11}{01} = (11)$. Thus *q* could be either true or false as expected. The inference could even proceed in the reverse direction, from *q* to *p*, but then we must compose from the right side of *R* to match the axes. Assume for instance *q* is true or $q = (10)^t$, then $\mathbf{p} = \mathbf{R} \circ \mathbf{q} = \binom{11}{01} \circ \binom{1}{0} = \binom{1}{0}$. Thus, if *q* is false and $p \Rightarrow q$, then *p* is false (*modus tollens*). The array based inference mechanism is even more general, because *R* can be any dimension *n*, $n > 0$, $n \in \mathbf{I}$. Given values of $n - 1$ variables, the possible outcomes of the remaining variable can be inferred by an *n*-dimensional inner product. Furthermore, given values of $n - d$ variables, $d \in \mathbf{I}$ and $0 < d < n$, then the truth-array connecting the remaining *d* variables can be inferred. Thus, using the fuzzy soft connectives various fuzzy soft inference rules can be developed.

## VII. An Application of Fuzzy Soft Relations to Decision Making Problems

The concept of fuzzy soft relation and its generalization can be used effectively for solving a wide range of Decision Making Problems. Using the fuzzy soft relation there is an inherent reduction in the computational effort. This fact is illustrated by several real life applications considered in this section. We consider here six real life applications viz., House Acquisition Problem, Job Allocation Problem, Investment Portfolio Problem, Fund Sources Problem, Manpower Recruitment Problem and Product Marketing Problem and show how the fuzzy soft relation can be used to generate effective solutions with least possible efforts. In all the applications, the membership values of the fuzzy soft sets are determined by considering the parameter set *E*, which are simulated by using MATLAB. Generally, two important membership functions Trapezoidal and Triangular membership functions are used for all cases. The final decision result changes if different membership values are given. The advantages of the fuzzy soft relations are also illustrated by comparing with other methods viz., probability and possibility distributions. The different values of the probability and possibility distributions are obtained through various real life simulations.

### A. Application 1: House Acquisition Problem

Using fuzzy soft relation we solve a variation of the House Acquisition Problem which was solved earlier by Maji et al [22]. Let $U = \{h_1, h_2, h_3, h_4, h_5, h_6, h_7\}$ be a set of seven houses and $E$ = {*expensive*, *wooden*, *beautiful*, *cheap*, *in green surroundings*, *concrete*, *moderately beautiful*, *by the roadside*} be the set of parameters. Let $(F_1, A_1)$ be the fuzzy soft set which describes the cost of the houses given by $(F_1, A_1) = \{F_1 (cheap) = \{h_1/ 1, h_2/ 0, h_3/ 1, h_4/ 0.2, h_5/ 1, h_6/ 0.2, h_7/ 1\}$, $F_1 (expensive) = \{h_1/ 0, h_2/ 1, h_3/ 0.1, h_4/ 0.9, h_5/ 0.3, h_6/ 1, h_7/ 0.7\}\}$. Let $(F_2, A_2)$ be the fuzzy soft set which describes the attractiveness of the houses given by $(F_2, A_2) = \{F_2 (beautiful) = \{h_1/ 1, h_2/ 0.4, h_3/ 1, h_4/ 0.4, h_5/ 0.6, h_6/ 0.8, h_7/ 0.7\}$, $F_2 (moderately\ beautiful) = \{h_1/ 0.3, h_2/ 0.7, h_3/ 0.5, h_4/ 0.6, h_5/ 0.2, h_6/ 0.3, h_7/ 0.4\}\}$. Let $(F_3, A_3)$ be the fuzzy soft set which describes the physical trait of the houses given by $(F_3, A_3) = \{F_3 (wooden) = \{h_1/ 0.2, h_2/ 0.3, h_3/ 1, h_4/ 1, h_5/ 1, h_6/ 0, h_7/ 1\}$, $F_3 (concrete) = \{h_1/ 0.7, h_2/ 0.9, h_3/ 0, h_4/ 0.1, h_5/ 0.3, h_6/ 0.8, h_7/ 0.6\}\}$. Similarly, $(F_4, A_4)$ be the fuzzy soft set which describes the characteristics of the place where the houses are located given by $(F_4, A_4) = \{F_4 (in\ green\ surroundings) = \{h_1/ 1, h_2/ 0.1, h_3/ 0.5, h_4/ 0.3, h_5/ 0.2, h_6/ 0.3, h_7/ 1\}$, $F_2 (near\ the\ roadside) = \{h_1/ 0.2, h_2/ 0.7, h_3/ 0.8, h_4/ 1, h_5/ 0.5, h_6/ 0.9, h_7/ 0.6\}\}$.

Suppose that Mr. Jones is interested in buying a house on the basis of his choice of parameters *beautiful*, *wooden*, *cheap*, *in green surroundings*. This implies that from the houses available in *U*, he should select the house that satisfies with all the parameters of his choice. The problem can be solved by virtue of the definition 7, a fuzzy soft relation $(R, C)$ among the fuzzy soft sets $(F_1, A_1)$, $(F_2, A_2)$, $(F_3, A_3)$ and $(F_4, A_4)$ of the houses of *U* which are *cheap*, *beautiful*, *wooden*, *in* *green surroundings*. By definition of the fuzzy soft relation $(R, C)$ is given by $(R, C) = \{R$ (*cheap, beautiful, wooden, in green surroundings*) = $\{h_1/ 0.2, h_2/ 0, h_3/ 0.5, h_4/ 0.2, h_5/ 0.2, h_6/ 0, h_7/ 0.7\}$. Thus, the house which best satisfies the requirement of Mr. Jones's choice is the house, which has the largest membership value in the relation. Here, $h_7$ has the largest membership value equal to 0.7; hence Mr. Jones will buy the house $h_7$. It is noted that the solution of the above problem obtained by Maji et al [21], [22] requires calculating the row sum, column sum and membership score for each house. This requires more computational time compared to the solution obtained by using fuzzy soft relation. So the method above is more efficient and economical.

**B. Application 2: Job Allocation Problem**

We now consider another decision-making problem of allocating a particular job to the best possible person who fulfills the requirements of the job. The problem is adopted from the Job Allocation Problem in Indian Industrial scenario. Let $U = \{p_1, p_2, p_3, p_4, p_5, p_6\}$ be the crisp set of six persons for the job. Let $E = \{$*enterprising, average, confident, confused, willing to take risks, unwilling to take risks*$\}$ be the set of parameters. Let $(F_1, A_1) = \{F_1$ (*enterprising*) = $\{p_1/ 0.5, p_2/ 0.7, p_3/ 0.3, p_4/ 0.1, p_5/ 0.8, p_6/ 0.9\}$, $F_1$ (*average*) = $\{p_1/ 0.3, p_2/ 0.1, p_3/ 0.5, p_4/ 0.8, p_5/ 0.05, p_6/ 0.7\}$ be the soft set describing the enterprising qualities of the person. Again $(F_2, A_2) = \{F_2$ (*confident*) = $\{p_1/ 0.6, p_2/ 0.8, p_3/ 0.5, p_4/ 0.2, p_5/ 0.9, p_6/ 0.8\}$, $F_3$ (*confused*) = $\{p_1/ 0.3, p_2/ 0.1, p_3/ 0.7, p_4/ 0.9, p_5/ 0.5, p_6/ 0.6\}$ be the soft set describing the confidence level of the person. Similarly, $(F_3, A_3) = \{F_3$ (*willing to take risks*) = $\{p_1/ 0.7, p_2/ 0.8, p_3/ 0.5, p_4/ 0.2, p_5/ 0.6, p_6/ 0.5\}$, $F_1$ (*unwilling to take risks*) = $\{p_1/ 0.3, p_2/ 0.07, p_3/ 0.65, p_4/ 0.95, p_5/ 0.1, p_6/ 0.6\}$ be the soft set describing the willingness level of the person.

Let us assume that the particular job requires an *enterprising*, *confident* person who is *willing to take risks*. Our problem is to find the candidate who best suits the requirements of the job. To solve this problem we use the definition 7, a fuzzy soft relation $(R, C)$ of the fuzzy soft sets $(F_1, A_1), (F_2, A_2), (F_3, A_3)$ of all candidates who are *enterprising*, *confident*, *willing to take risks*. By definition, $(R, C)$ is given by $(R, C) = \{p_1/ 0.21, p_2/ 0.45, p_3/ 0.08, p_4/ 0.05, p_5/ 0.43, p_6/ 0.36\}$. From the relation it is evident that the most suitable candidate for the job is $p_2$ who possesses the greatest membership value in the relation $(R, C)$.

Now, we present the probability and possibility distributions for the above problem corresponding to one specific parameter. Considering the risk taking parameter viz., *willing to take risks* from the parameter set $E$ we have the following probability distribution *prob* for the persons $p_i$; $i = 1,\ldots\ldots,6$ from the set $U$.

| $p_i$ | 1 | 2 | 3 | 4 | 5 | 6 |
|---|---|---|---|---|---|---|
| *prob* ($p_i$) | 0.25 | 0.55 | 0.1 | 0.1 | 0 | 0 |

**Table 8: Probability values of person $p_i$ with respect to risk taking parameter**

Again, a fuzzy set expressing the risk taking attitude of the persons $p_i$; $i = 1,\ldots\ldots,6$ from the set $U$ may be the expressed using the following possibility distribution $\pi$,

| $p_i$ | 1 | 2 | 3 | 4 | 5 | 6 |
|---|---|---|---|---|---|---|
| $\pi(p_i)$ | 1 | 1 | 1 | 1 | 0.8 | 0.7 |

**Table 9: Possibility values of person $p_i$ with respect to risk taking attitude**

It is to be noted that each possibility is at least as high as the corresponding probability. Further, the sum of *prob* ($p_i$) is always equal to 1 but $\pi(p_i)$ may be equal to, greater or less than 1. As it is obvious from the above discussion that it is much easier to represent a large amount of information viz., different parameters using the fuzzy soft relations which is the prime requirement in most decision making situations, because the final decision to the problem is dependent on various associated parameters. Besides this, using probability and possibility distributions only a partial representation of the information is possible, which leads to final decision results which are inaccurate and incomplete. Finally, the solution to the problem is obtained with minimal computational effort using fuzzy soft relations.

## C. Application 3: Investment Portfolio Problem

The Investment Portfolio Problem is simulated from ICICI Prudential Financial Services, India. Let $U = \{i_1, i_2, i_3, i_4, i_5, i_6\}$ be a set of six investments at the disposal of the investor to invest some money and $E = \{$*investment price, advance mobilization, period, returns, risk, security*$\}$ be the set of parameters. Let ($F_1$, $A_1$) be the fuzzy soft set which describes the attractiveness of investments to the customers given by ($F_1$, $A_1$) = {$F_1$ (*investment price*) = {$i_1$/ 0.1, $i_2$/ 0.7, $i_3$/ 0.4, $i_4$/ 0.9, $i_5$/ 0.6, $i_6$/ 0.5}, $F_1$ (*advance mobilization*) = {$i_1$/ 0.7, $i_2$/ 0.1, $i_3$/ 1, $i_4$/ 0.8, $i_5$/ 0.4, $i_6$/ 0.5}. Let ($F_2$, $A_2$) be the fuzzy soft set which describes the rate of returns on the investments given by ($F_2$, $A_2$) = {$F_2$ (*period*) = {$i_1$/ 0.5, $i_2$/ 0.6, $i_3$/ 0.5, $i_4$/ 0.8, $i_5$/ 0.7, $i_6$/ 1}, $F_2$ (*high returns*) = {$i_1$/ 0.9, $i_2$/ 0.6, $i_3$/ 0.3, $i_4$/ 1, $i_5$/ 0.7, $i_6$/ 0.8}. Let ($F_3$, $A_3$) be the fuzzy soft set which describes the risk factor of the investments given by ($F_3$, $A_3$) = {$F_3$ (*risk*) = {$i_1$/ 0.9, $i_2$/ 0.8, $i_3$/ 0.7, $i_4$/ 0.6, $i_5$/ 1, $i_6$/ 0.5}, $F_3$ (*security*) = {$i_1$/ 0.4, $i_2$/ 0.7, $i_3$/ 0.2, $i_4$/ 0.3, $i_5$/ 1, $i_6$/ 0.9}.

Let us consider that the person wishes to have an investment which has *advance mobilization* gives *high returns* and is of *secured* nature. The problem involves in finding an investment which maximum returns to the person. To solve this problem we use the definition 7, a fuzzy soft relation ($R$, $C$) of the fuzzy soft sets ($F_1$, $A_1$), ($F_2$, $A_2$), ($F_3$, $A_3$) of all investments which are *advance mobilized* gives *high returns* and is of *secured* nature. By definition, ($R$, $C$) is given by ($R$, $C$) = {$i_1$/ 0.25, $i_2$/ 0.04, $i_3$/ 0.06, $i_4$/ 0.24, $i_5$/ 0.28, $i_6$/ 0.36}. From the relation it is obvious that the most profitable investment for the person is $i_6$ which has the greatest membership value in the relation ($R$, $C$). Considering the investment parameter viz., *advance mobilization* from the parameter set $E$ we have the following probability distribution *prob* for the investment $i_k$; $k = 1,\ldots\ldots, 6$ from the set $U$.

| $i_k$ | 1 | 2 | 3 | 4 | 5 | 6 |
|---|---|---|---|---|---|---|
| *prob* ($i_k$) | 0.19 | 0.36 | 0.2 | 0.1 | 0.1 | 0.05 |

**Table 10: Probability values of investment $i_k$ with respect to advance mobilization**

Again, a fuzzy set expressing the *advance mobilization* of the investment $i_k$; $k = 1,\ldots\ldots,6$ from the set $U$ may be the expressed using the following possibility distribution $\pi$,

| $i_k$ | 1 | 2 | 3 | 4 | 5 | 6 |
|---|---|---|---|---|---|---|
| $\pi$ ($i_k$) | 0.36 | 0.69 | 1 | 0.54 | 1 | 0.86 |

**Table 11: Possibility values of investment $i_k$ with respect to advance mobilization**

As evident from the above discussion that a large amount of information is easily represented using the fuzzy soft relations as compared to other methods which leads to much more precise and accurate decision results. Also, the computational effort is minimized using fuzzy soft relations.

### D. Application 4: Fund Sources Problem

The Fund Sources Problem is taken from Axis Bank, India. Let $U = \{s_1, s_2, s_3, s_4, s_5\}$ be a set of five fund sources available for a Manager in a Banking System and $E = \{$*term deposit*, *demand deposit*, *fund pricing*, *fund mobility*, *liquidity*, *investment*$\}$ be the set of parameters. Let $(F_1, A_1)$ be the fuzzy soft set which describes the maturity pattern of deposit given by $(F_1, A_1) = \{F_1$ (*term deposit*) $= \{s_1/ 0.95, s_2/ 0.86, s_3/ 0.79, s_4/ 1, s_5/ 0.21\}$, $F_1$ (*demand deposit*) $= \{s_1/ 0.75, s_2/ 0.69, s_3/ 0.58, s_4/ 0.46, s_5/ 0.29\}$. Let $(F_2, A_2)$ be the fuzzy soft set which describes the competition in the fund market given by $(F_2, A_2) = \{F_2$ (*fund pricing*) $= \{s_1/ 0.15, s_2/ 0.27, s_3/ 0.37, s_4/ 0.78, s_5/ 0.35\}$, $F_2$ (*fund mobility*) $= \{s_1/ 0.5, s_2/ 0.66, s_3/ 0.7, s_4/ 0.19, s_5/ 1\}$. Let $(F_3, A_3)$ be the fuzzy soft set which describes the strength of the organization to fulfill the commitment given by $(F_3, A_3) = \{F_3$ (*liquidity*) $= \{s_1/ 1, s_2/ 0.75, s_3/ 0.53, s_4/ 0.48, s_5/ 0.96\}$, $F_3$ (*investment*) $= \{s_1/ 0.24, s_2/ 0.39, s_3/ 0.85, s_4/ 1, s_5/ 0.44\}$.

Let us assume the Manager Mr. Steve intends to have a fund source which possesses the attributes such as *term deposit*, *fund mobility* and *liquidity*. This means that from the fund sources in $U$, he must select the fund source that satisfies with all the parameters of his requirements. To solve this problem we use the definition 7, a fuzzy soft relation $(R, C)$ of the fuzzy soft sets $(F_1, A_1)$, $(F_2, A_2)$, $(F_3, A_3)$ of all fund sources which has the attributes *term deposit*, *fund mobility* and *liquidity*. By definition, $(R, C)$ is given by $(R, C) = \{s_1/ 0.48, s_2/ 0.43, s_3/ 0.24, s_4/ 0.09, s_5/ 0.20\}$. Thus, the fund source which best satisfies the requirement of the Manager Mr. Steve's choice is the fund source, which has the largest membership value in the relation. Here, $s_1$ has the largest membership value equal to 0.48; hence Mr. Steve's will choose the fund source $s_1$.

Considering the *fund mobility* parameter from the set $E$ we have the following probability distribution *prob* for the fund source $m_k$; $k = 1,\ldots\ldots, 5$ from the set $U$.

| $m_k$ | 1 | 2 | 3 | 4 | 5 |
|---|---|---|---|---|---|
| *prob* $(m_k)$ | 0.35 | 0.21 | 0.04 | 0.2 | 0.2 |

Table 12: Probability values of fund source $m_k$ with respect to fund mobility

Again, a fuzzy set expressing the *fund mobility* of the fund source $m_k$; $k = 1,\ldots\ldots,5$ from the set $U$ may be the expressed using the following possibility distribution $\pi$,

| $m_k$ | 1 | 2 | 3 | 4 | 5 |
|---|---|---|---|---|---|
| $\pi (m_k)$ | 1 | 1 | 1 | 1 | 0.46 |

Table 13: Possibility values of fund source $m_k$ with respect to fund mobility

Form the above discussion it is evident that the fuzzy soft relation represents voluminous information easily as compared to other methods from which more precise and less vague decision results are obtained. The computational effort required is less.

### E. Application 5: Manpower Recruitment Problem

The Manpower Recruitment Problem is adopted from Tata Consultancy Services, India. Let $U = \{m_1, m_2, m_3, m_4, m_5, m_6, m_7\}$ be a set of seven programmers to be recruited by a Software Development Organization by the Human Resources Manager and $E = \{$*hardworking, disciplined, honest, obedient, intelligence, innovative, entrepreneurial attitude, aspirant*$\}$ be the set of parameters. Let $(F_1, A_1)$ be the fuzzy soft set which describes the punctuality of the programmer given by $(F_1, A_1) = \{F_1$ (*hardworking*) $= \{m_1/ 0.17, m_2/ 1, m_3/ 0.88, m_4/ 0.26, m_5/ 0.55, m_6/ 0.28, m_7/ 0.98\}$, $F_1$ (*disciplined*) $= \{m_1/ 1, m_2/ 0.33, m_3/ 0.7, m_4/ 0.64, m_5/ 0.4, m_6/ 0.3, m_7/ 0.57\}$. Let $(F_2, A_2)$ be the fuzzy soft set which describes the truth in the behavior of the programmer given by $(F_2, A_2) = \{F_2$ (*honest*) $= \{m_1/ 0.09, m_2/ 0.81, m_3/ 0.05, m_4/ 1, m_5/ 0.45, m_6/ 0.24, m_7/ 0.18\}$, $F_2$ (*obedient*) $= \{m_1/ 1, m_2/ 0.56, m_3/ 1, m_4/ 0.04, m_5/ 0.65, m_6/ 0.97, m_7/ 1\}$. Let $(F_3, A_3)$ be the fuzzy soft set which describes the innovativeness in the programmer's attitude given by $(F_3, A_3) = \{F_3$ (*intelligence*) $= \{m_1/ 0.13, m_2/ 0.93, m_3/ 0.08, m_4/ 0.36, m_5/ 1, m_6/ 0.48, m_7/ 0.47\}$, $F_3$ (*innovative*) $= \{m_1/ 0.54, m_2/ 0.22, m_3/ 0.16, m_4/ 0.42, m_5/ 0.5, m_6/ 0.2, m_7/ 0.99\}$. Let $(F_4, A_4)$ be the fuzzy soft set which describes the exploratory attitude of the programmer given by $(F_4, A_4) = \{F_4$ (*entrepreneurial attitude*) $= \{m_1/ 1, m_2/ 0.72, m_3/ 0.7, m_4/ 0.64, m_5/ 0.7, m_6/ 0.8, m_7/ 0.65\}$, $F_4$ (*aspirant*) $= \{m_1/ 0.14, m_2/ 0.3, m_3/ 0.82, m_4/ 0.62, m_5/ 1, m_6/ 0.05, m_7/ 0.77\}$.

Let us assume that the Human Resources Manager Mr. Adams wants to recruit a programmer who has the qualities like *hardworking*, *honest*, *innovative* and *entrepreneurial attitude*. Thus, from the available candidates in $U$, he should select the programmer who satisfies with all the parameters of his requirements. To solve this problem we use the definition 7, a fuzzy soft relation $(R, C)$ of the fuzzy soft sets $(F_1, A_1)$, $(F_2, A_2)$, $(F_3, A_3)$, $(F_4, A_4)$ of all the programmers who have the qualities *hardworking*, *honest*, *innovative* and *entrepreneurial attitude*. By definition, $(R, C)$ is given by $(R, C) = \{m_1/ 0.01, m_2/ 0.13, m_3/ 0.05, m_4/ 0.07, m_5/0.09, m_6/0.01, m_7/0.11\}$. Thus, from the above calculations we infer that the second programmer, $m_2$ has the largest membership value i.e., 0.13 in the relation; hence the Human Resource Manager will select the second programmer for the Software Development job.

Considering the *innovative* parameter from the parameter set $E$ we have the following probability distribution *prob* for the manpower recruitment $r_i$; $i = 1,\ldots\ldots,7$ from the set $U$.

| $r_i$ | 1 | 2 | 3 | 4 | 5 | 6 | 7 |
|---|---|---|---|---|---|---|---|
| *prob* $(r_i)$ | 0.19 | 0.16 | 0.15 | 0.15 | 0.05 | 0.16 | 0.14 |

**Table 14: Probability values of recruitment $r_i$ with respect to innovativeness parameter**

Again, a fuzzy set expressing the innovativeness parameter for the manpower recruitment $r_i$; $i = 1,\ldots\ldots,7$ from the set $U$ may be the expressed using the following possibility distribution $\pi$,

| $r_i$ | 1 | 2 | 3 | 4 | 5 | 6 | 7 |
|---|---|---|---|---|---|---|---|
| $\pi(r_i)$ | 1 | 1 | 1 | 0.55 | 1 | 0.69 | 1 |

**Table 15: Possibility values of recruitment $r_i$ with respect to innovativeness parameter**

Form the above discussion it is evident that the fuzzy soft relation represents voluminous information easily as compared to other methods from which more precise and less vague decision results are obtained. The computational effort required is less.

**F. Application 6: Product Marketing Problem**

The Product Marketing Problem is simulated from Khosla Electronics, Kolkata, India. Let $U = \{t_1, t_2, t_3, t_4, t_5, t_6, t_7\}$ be a set of six brands of televisions to be sold in an international market by a retail outlet owner and $E = \{$*price*, *modern technology*, *portability*, *screen size*, *weight*, *longevity*, *picture clarity*, *audible sound*$\}$ be the set of parameters. Let $(F_1, A_1)$ be the fuzzy soft set which describes the price effectiveness of the television given by $(F_1, A_1) = \{F_1$ (*price*) $= \{t_1/ 0.6, t_2/ 0.5, t_3/ 0.56, t_4/ 1, t_5/ 0.01, t_6/ 0, t_7/ 0.99\}$, $F_1$ (*modern technology*) $= \{t_1/ 1, t_2/ 0.75, t_3/ 0.43, t_4/ 0.33, t_5/ 1, t_6/ 0.83, t_7/ 0.04\}$. Let $(F_2, A_2)$ be the fuzzy soft set which describes the television's lightness aspect given by $(F_2, A_2) = \{F_2$ (*portability*) $= \{t_1/ 0.06, t_2/ 0.7, t_3/ 1, t_4/ 0.05, t_5/ 0, t_6/ 1, t_7/ 0.8\}$, $F_2$ (*weight*) $= \{t_1/ 1, t_2/ 0.87, t_3/ 0.03, t_4/ 0.23, t_5/ 0.16, t_6/ 0.75, t_7/ 1\}$. Let $(F_3$,

$A_3$) be the fuzzy soft set which describes the dimensionality of the television given by ($F_3$, $A_3$) = {$F_3$ (*screen size*) = {$t_1$/ 0.61, $t_2$/ 0.1, $t_3$/ 0.2, $t_4$/ 0.25, $t_5$/ 0.67, $t_6$/ 0.05, $t_7$/ 1}, $F_3$ (*audible sound*) = {$t_1$/ 0.83, $t_2$/ 1, $t_3$/ 0.21, $t_4$/ 0.45, $t_5$/ 0, $t_6$/ 0.74, $t_7$/ 0.84}. Let ($F_4$, $A_4$) be the fuzzy soft set which describes the durability of the television given by ($F_4$, $A_4$) = {$F_4$ (*longevity*) = {$t_1$/ 1, $t_2$/ 0.4, $t_3$/ 0.7, $t_4$/ 0.55, $t_5$/ 1, $t_6$/ 0.91, $t_7$/ 0.97}, $F_4$ (*picture clarity*) = {$t_1$/ 0.12, $t_2$/ 0.89, $t_3$/ 0.39, $t_4$/ 1, $t_5$/ 0.6, $t_6$/ 0, $t_7$/ 0.46}.

Let us assume that the retail owner wants to maximize his profits by selling the television brand which possesses the attributes such as *modern technology*, *portability*, *audible sound* and *picture clarity*. Hence from the available brands of the television sets in *U*, he must select the television brand which satisfies with all the parameters of the requirements. To solve this problem we use the definition 7, a fuzzy soft relation (*R*, *C*) of the fuzzy soft sets ($F_1$, $A_1$), ($F_2$, $A_2$), ($F_3$, $A_3$), ($F_4$, $A_4$) of all the brands of the television sets which has the attributes *modern technology*, *portability*, *audible sound* and *picture clarity*. By definition, (*R*, *C*) is given by (*R*, *C*) = {$t_1$/ 0.01, $t_2$/ 0.46, $t_3$/ 0.04, $t_4$/ 0.08, $t_5$/ 0, $t_6$/ 0, $t_7$/ 0.01}.

Thus, the television brand which best meets the requirement of the retail owner is the television brand, which has the maximum membership value in the relation. Here, $t_2$ has the largest membership value equal to 0.46; hence retail outlet owner will consider the television brand $t_2$ to maximize his profits.

Considering the *price* parameter from the parameter set *E* we have the following probability distribution *prob* for the product $t_i$; $i = 1,\ldots\ldots,7$ from the set *U*.

| $t_i$ | 1 | 2 | 3 | 4 | 5 | 6 | 7 |
|---|---|---|---|---|---|---|---|
| *prob* ($t_i$) | 0.16 | 0.04 | 0.1 | 0.1 | 0.15 | 0.3 | 0.15 |

Table 16: Probability values of product $t_i$ with respect to price parameter

Again, a fuzzy set expressing the price parameter for the product $t_i$; $i = 1,\ldots\ldots,7$ from the set *U* may be the expressed using the following possibility distribution $\pi$,

| $t_i$ | 1 | 2 | 3 | 4 | 5 | 6 | 7 |
|---|---|---|---|---|---|---|---|
| $\pi$ ($t_i$) | 1 | 0.89 | 1 | 1 | 1 | 0.96 | 1 |

Table 17: Possibility values of product $t_i$ with respect to price parameter

Form the above discussion it is evident that the fuzzy soft relation represents voluminous information easily as compared to other methods from which more precise and less vague decision results are obtained. The computational effort required is less.

## VIII. Conclusion

This Paper presents the concepts of Soft Relation and Fuzzy Soft Relation to solve various Decision Making Problems in the Engineering, Management, and Social Science domains. These problems often involve data that are imprecise, uncertain and vague nature. A number of solutions have been proposed for such problems using Probability Theory, Fuzzy Set Theory, Rough Set Theory, Vague Set Theory, Approximate Reasoning Theory etc in the past. These techniques however lack in parameterization of the tools due to which they could not be applied successfully in dealing with such problems. The Soft Set and Fuzzy Soft Set concepts possess certain parameterization features which are certain extensions of crisp and fuzzy relations respectively and have a rich potential for application to the Decision Making Problems. This fact is evident from the theoretical analysis which illustrates the rationality of the proposed method. Finally, we make use of these concepts in solving some real life Decision Making Problems and present the advantages of the Fuzzy Soft Sets compared to other paradigms.

## References

[1] Atanassov K., Intuitionistic Fuzzy Sets, *Fuzzy Sets and Systems*, 1986, 20: 87 – 96.

[2] Atanassov K., Operators over interval valued Intuitionistic Fuzzy Sets, *Fuzzy Sets and Systems*, 1994, 64: 159 – 174.

[3] Baker J.F., Stewart T. S., Long C. R. & Cartwright T. C., Multiple Regressions and Principal Components Analysis of Puberty and Growth in Cattle, *Journal of Animal Science,* 1988, 66: 2147 - 2158.

[4] Brehmer B., The development of social judgment theory, 1988, New York: Wiley.

[5] Brunswik, E., Systematic and representative design of psychological experiments, with results in physical and social perception, 1947.

[6] Craiger P. & Coovert. M. D., Modeling dynamic social and psychological processes with fuzzy cognitive maps, In *Proceedings of. 3rd IEEE Conference on Fuzzy Systems*, 1994.

[7] Ganzach Y. & Schul Y., The influence of quantity of information and valence framing on decision. Acta Psychologica, 1995, 89: 23-36.

[8] Gau W. L. & Buchrer D. J., Vague Sets, *IEEE Transactions System*, *Man*, *Cybernetics*, 1993, 23: 610 – 614.

[9] Gobet F. & Ritter F. E., Individual Data Analysis and Unified Theories of Cognition: A methodological proposal, *Proceedings of the 3rdInternational Conference on Cognitive Modeling*, Veenendaal (NL), Universal Press, 2000, 150 - 157.

[10] Gorzalzany M. B., A method of inference in approximate reasoning based on interval valued Fuzzy Sets, *Fuzzy Sets and Systems*, 1987, 21: 1 – 17.

[11] Gujon R. M., Assessment, measurement, and prediction for personnel decisions, Wiley InterScience, 1998.

[12] Hammond K.R., Introduction to Brunswikian theory and methods: New Directions for Methodology of Social and Behavioral Science, 1980, 3: 1 - 11.

[13] Hesketh, T. & Hesketh, B., Computerized Fuzzy Ratings: The concept of a fuzzy class. Behavior Research Methods, Instruments and Computers, 1994, 26: 272 – 277.

[14] James Reason, *Human Error*. Ashgate, 1990.

[15] Jang J. R. & Sun C., Neuro-Fuzzy Modeling and Control, *Proceedings of the IEEE*, 1995, 83: 378 - 406.

[16] Jang S., Sun C.T. & Mizutani E., Neuro-Fuzzy and Soft Computing, Prentice Hall, 1997.

[17] Kahneman D. & Tversky A., Choice, Values, Frames, The Cambridge University Press, 2000.

[18] Kapoor V. K., Operations Research – Techniques for Management, 2006.

[19] KIir G. & Yuan. B., Fuzzy Sets and Fuzzy Logic: Theory and Applications, Prentice Hall, 1995.

[20] Kosko B., Neural Networks and Fuzzy Systems, Engelwood Cliffs, Prentice Hall, New Jersey, 1992.

[21] Maji P. K., Biswas R. & Roy A. R., On Soft Set Theory, *Computers and Mathematics with Applications*, 2003, 41: 1 – 8.

[22] Maji P. K., Biswas R. & Roy A. R., Fuzzy Soft Sets, *The Journal of Fuzzy Mathematics*, 2001, 9(3): 589 – 602.

[23] Martinsons M. G., Comparing the Decision Styles of American, Chinese and Japanese Business Leaders, *Proceedings of Academy of Management Meetings*, Washington, DC, 2001.

[24] Myers I. B. & Myers I., Introduction to Type: A description of the theory and applications of the Myers-Briggs type indicator, Consulting Psychologists Press, Palo Alto California, 1962.

[25] Molodostov D., Soft Set Theory – First results, *Computers, Mathematics with Applications*, 1999, 37: 19 – 31.

[26] Newell Allen & Herbert Simon., Human Problem Solving. Englewood Cliffs, N.J.: Prentice-Hall, 1972.

[27] Paul I. Barton P.I., Optimization of Hybrid Systems, Computers & Chemical Engineering, Science Direct, 2006, 30(11): 1576-1589.

[28] Pawlak Z., Rough Sets, *International Journal of Information & Computer Science*, 1999, 11: 341 – 356.

[29] Pawlak Z., Hard Sets and Soft Sets, *ICS Research Report*, *Institute of Computer Science*, Poland, 1994.

[30] Thielle H., On the concepts of Qualitative Fuzzy Sets, *IEEE International Symposium on Multi-valued Logic*, Japan, 1999.

[31] Thurstone L. L., *The Nature of Intelligence*, London, Routledge, 1973.

[32] Yao Y. Y., Relational interpretations of neighborhood operators and Rough Set approximation operators, *Information Science*, 1998, 111: 239 – 259.

[33] Zadeh L. A., Fuzzy Sets, *Information and Control*, 1965, 8:338 – 353.

[34] Zadeh L. A., Fuzzy Logic, Neural Networks, and Soft Computing, *Communications of the ACM*, 1994, 37(3): 77 – 84.

**About the Authors:**

**Arindam Chaudhuri** is a Lecturer in Mathematics and Computer Science at Meghnad Saha Institute of Technology, Kolkata, India. He holds Bachelors degree in Physics from Calcutta University, Master degree in Information Systems and Operations Research from All India Management Association, New Delhi, Master degree in Computer Applications from Networked University, Chattisgarh and Master of Technology in Computer Science Engineering from Doeacc Society, New Delhi. He is working towards PhD degree in Computer Science from Netaji Subhas Open University, Kolkata, India. His research interests include Soft Computing and Optimization Algorithms.

**Kajal De** is a Professor in Mathematics at Netaji Subhas Open University, Kolkata, India. She holds Bachelors, Masters and PhD degrees in Applied Mathematics from Calcutta University. She has over 15 years of teaching and research experience. She has published a number of papers in leading International Journals in Applied Mathematics. Her research interests include Fuzzy Sets and Optimization Algorithms.

**Dipak Chatterjee** is a Distinguished Professor in Mathematics at St. Xavier's College, Kolkata, India. He holds Bachelors, Masters and PhD degrees in Applied Mathematics from Calcutta University. He has over 30 years of teaching and research experience. He has published about 70 papers in leading International Journals and has authored about 25 books in Mathematics, Statistics and Computer Science. His research interests include Soft Computing and Optimization Algorithms.